\documentclass[fleqn]{article}

\usepackage{aaai}
\nocopyright

\newtheorem{example}{Example}

\title{Logic Programming for Describing and Solving Planning Problems}
\author{Maurice Bruynooghe\\
Department of Computer Science, \\
Katholieke Universiteit Leuven, \\
Celestijnenlaan 200A, B-3001 Heverlee, Belgium \\
email{\{maurice\}}@cs.kuleuven.ac.be,\\
}
\begin{document} 

\maketitle

\begin{abstract}
  A logic programming paradigm which expresses solutions to problems
  as stable models has recently been promoted as a declarative
  approach to solving various combinatorial and search problems,
  including planning problems. In this paradigm, all program rules are
  considered as constraints and solutions are stable models of the
  rule set. This is a rather radical departure from the standard
  paradigm of logic programming. In this paper we revisit abductive
  logic programming and argue that it allows a programming style which
  is as declarative as programming based on stable models. However,
  within abductive logic programming, one has two kinds of rules. On
  the one hand predicate definitions (which may depend on the
  abducibles) which are nothing else than standard logic programs
  (with their non-monotonic semantics when containing with negation);
  on the other hand rules which constrain the models for the
  abducibles.  In this sense abductive logic programming is a smooth
  extension of the standard paradigm of logic programming, not a
  radical departure.
\end{abstract}

{\bf keywords:} planning, abduction, non-monotonic reasoning.

\section{Introduction}
\label{sec:intro}

A number of recent papers argue in favour of a new logic programming
paradigm based on stable model semantics \cite{GL88}.  \cite{MT99}
introduces Stable Logic Programming as a novel programming paradigm.
The language is basically DATALOG extended with negation. Stable
models of programs form a finite family of finite sets, hence the
solutions to search problems can be represented as stable models of
Stable Logic Programs. The programming style is to introduce
constraints which restrict the stable models to the solutions.  The
paper refers to various related efforts; the closest and most
prominent one is that of \cite{Nie98,Nie00} which proposes function
free logic programming with stable model semantics as a paradigm for
constraint programming which ``brings advantages of logic programming
based knowledge representation techniques to constraint programming''.
The {\bf Smodels} system of \cite{NS96} has evolved in one of the
leading implementation efforts.  Lifschitz \cite{Lif99,Lif-ICLP99}
introduces Answer Set Programming and explores its use in planning.
The formalism is slightly more expressive: disjunctive heads are
allowed and two forms of negation are used. However simple
transformations can eliminate disjunctive heads and classical negation
and Answer Set Programming basically relies on the same
implementations as Stable Logic Programming for execution (for
computing stable models). In the rest of this paper, we refer to this
programming paradigm as {\em Stable Logic Programming}.

When the solution of a problem is some finite set, encoding it as a
term which is the computed answer of a query of a logic program
induces a level of indirection which increases the distance between
the problem domain and the program. Indeed, in this ``term-based''
programming style \cite{Danny99}, knowledge about the problem domain
is not a ``first class citizen'' in the problem representation. One
can rightfully say that such programs are less declarative and hence
worse from knowledge representation point of view than their
counterparts based on stable model semantics.  As a concrete example,
consider the n-queens problem.  The traditional (constraint) logic
programming approach represents the solution as a list of
column-positions with the position in the list encoding the row. This
is arguable less declarative than the solution in the above formalisms
where the stable model can be a set of facts {\tt position(i,j)}.

However the stable model paradigm is a rather drastic departure from
the standard paradigm of logic programming \cite{MT99}. Abduction
\cite{Kakas,Kakas98} is another formalism which also computes models (for the
so called {\em abducibles}). It can be interpreted as a smooth
extension of the traditional logic programming paradigm. Indeed, part
of an abductive logic program is a traditional logic program
consisting of predicate definitions. The extension consists of
additional rules that are constraining the possible models of the
abducibles. Consequently, a solution is not only a substitution for
the variables in the query but, more importantly, a model for the
abducibles. At a rather informal level, this paper revisits logic
programming, the treatment of negation, the extension to abduction and
discusses the suitability of abductive logic programming to solve the
problems for which Stable Logic Programming is being promoted, in
particular its use in solving planning problems.

The purpose of this paper is to argue that the abductive logic
programming paradigm is comparable to the stable model logic
programming paradigm in bringing advantages of better knowledge
representation techniques to constraint programming in general and to
planning problems in particular. Moreover, being an extension of the
standard logic programming paradigm instead of a radical departure,
abductive logic programming is perhaps easier to learn for experienced
logic programmers than stable logic programming.

\section{Logic programming revisited}
\label{sec:LP}

The impact of Kowalski's seminal paper \cite{Kow74} on the use of
definite Horn clauses for programming is in a large part due to the
combination of extreme simplicity of the formalism (definite Horn
clauses of the form $ h(t) \leftarrow b(t_1),\ldots,b(t_n)$ with
$h(t), b(t_1),\ldots,b(t_n)$ atoms) with high expressiveness (the
power of the universal Turing machine) and with the ability to control
the procedural behaviour. Being rooted in first order logic theorem
proving, definite Horn clauses were initially introduced as a subset
of first order logic. When it came to semantics, in particular
semantics of negation, it became gradually clear that logic programs
are not a subset of first order logic. For definite programs there is
a consensus that the least Herbrand model provides the appropriate
declarative meaning. It {\em defines} which atoms are true. Atoms
which are not true are false, hence the Closed World Assumption
\cite{Reiter78} is the correct semantics for deriving negative
information from definite programs. Although negation as finite
failure and completion semantics \cite{Clark78} characterises what SLD
can proof about definite programs, it is not the appropriate
semantics. The Closed World semantics has the practical drawback that
the set of false atoms is in general not recursively enumerable, hence
not effectively computable \cite{Apt90}.  (The lack of decidability is
a major motivation for the work on termination analysis of logic
programs.) Circumscribing the unique least model requires higher order
logic and definite programs are abbreviations for such higher
order theories \cite{Marc98}. Negation is {\em classical} negation with
respect to the implicit higher order theory.

The situation is less clear once general clauses (with negation in the
body) are considered. There is an overwhelming amount of literature
about different semantics, different extensions, \ldots.  Having
rejected completion semantics, one major direction is to give up the
idea of a unique model. It leads to the stable model semantics and the
programming paradigms mentioned in the introduction (weaker than the
universal Turing machine, because giving up functions). The other
direction sticks to the unique model property: stratification, local
stratification, \ldots, well-founded semantics \cite{Gelder91}.
Although one model is but a special case of several models, I have a
preference for the unique model approach. My argument is that when
performing a {\em programming} task, the programmer should define a
unique model for the task at hand. Moreover, that unique model should
be complete, in other words, the well-founded model should be two
valued. If not, the program has an error. In this view, a logic
program is a set of inductive definitions (in the mathematical sense)
defining a unique model. Note that, once a concept\footnote{One can
  consider each ground instance of an atom as a separate concept.} is
defined, its negation can be used to define another concept. This
insight is at the basis of the various forms of stratification. As we
argued for the case of CWA and definite programs, such use of negation
is classical negation, not negation as failure to proof, with respect
to the implicit higher order theory which is associated with the
program. Denecker \cite{Marc98} has investigated in depth the
mathematical notion of inductive definition and has shown that general
programs with a two valued well-founded semantics are correct
inductive definitions. Note that normal programs are non-monotonic.
Extending a normal program with extra clauses can cause non-monotonic
changes in the model of the implicit higher order theory.

A problem is that not all programming tasks can be achieved by writing
an inductive definition. Coming back to the n-queens problem, it is
impossible ---apart from giving all solutions as facts--- to give an
inductive definition of the {\tt position/2} predicate that describes
the positions of the n-queens on the board. The solution traditionally
pursued in (constraint) logic programming is to encode the set of true
{\tt position/2} atoms in a data structure and to write a correct
definition of a program searching the space of possible configurations
for a safe one.  A typical program is as follows:
\begin{verbatim}
queens(Q,N) :- generate(Q,N,N),  
               safe(Q),  
               instantiate(Q).
generate([],0,_).
generate([X|T],M,N) :- M > 0,  
                       X in 1..N,  
                       M1 is M - 1,  
                       generate(T,M1,N). 
safe([]). 
safe([X|T]) :- noAttack(X,1,T),
               safe(T). 
noAttack(_,_,[]). 
noAttack(X,N,[Y|Z]) :- X \= Y,  
                       Y \= X + N,  
                       X \= Y + N,  
                       S is N + 1,
                       noAttack(X,S,Z). 
instantiate([]).
instantiate([X|T]) :- enum(X),  
                      instantiate(T).
\end{verbatim}
This program is written in a rather procedural style with a double
recursion inside the {\tt safe/2} predicate to set up the constraints.
This is quite a bit away of the ideal of declarative programming. The
mapping between the program representation and the problem domain is
not as simple as it should be in truly declarative programming.

Here Abductive Logic Programming comes to the rescue.  Our approach is
to distinguish between on one hand predicates one does not know ({\em
  abducible predicates}) and on the other hand predicates one knows
and is able to define correctly and completely ({\em defined
  predicates}) when assuming a correct and complete definition of the
abducible ones.  Although one is unable to define the abducible
predicates, one typically has some partial knowledge about them. In
the queens problem, not every collection of {\tt position/2} atoms
makes up a valid definition. They should satisfy the constraint that
they do not attack each other and that there is one on each row.
Hence, besides a definition component, a program should also have a
constraint component. Then, given a set of abducible predicates $A$, a
definition $D$ and a set of constraints $C$, the task of an abductive
solver is to come up with a definition $\Delta$ (as a set of facts) of
the abducible predicates $A$ such that the constraints are true for
the definition $D \cup \Delta$, i.e.\ $D \cup \Delta \models C$. Note
that we impose that $D \cup \Delta$ is a correct inductive definition;
hence its unique model includes $\Delta$.

To make things concrete, let us look at a program for the n-queens
problem.

\begin{example}{n-queens}.
\label{ex:queens}
\begin{verbatim}
abducible position/2.

% DEFINITIONS
size(8).
row(R) :- size(N), R in 1..N.
column(C) :- size(N), C in 1..N.
row_has_queen(R) :- position(R,C).

% CONSTRAINTS
% the arguments of position/2 have the 
%               correct types
row(R)  <- position(R,C).
column(C)  <- position(R,C).
% at least one queen on each row
row_has_queen(R) <- row(R).
% no more than one queen on each row
C1=C2 <- position(R,C1), position(R,C2).
% the configuration is illegal if 
% a queen attacks a queen on a higher row
false <- position(R1,C1), position(R2,C2), 
         R1<R2,
         (C1=C2 ; abs(R2-R1)=abs(C2-C1)).
\end{verbatim}
\end{example}

The program has three components. The first component declares which
predicates are the abducible ones. The second component consists of
the definitions. We follow a PROLOG-like notation for it. It has a
definition for the size of the board, for the concepts row and column,
and for the concept of a row which has a queen. Some of the
definitions make use of an assumed ``built-in'' {\tt X in Y..Z} which
returns the integers $X$ in the interval $Y$ to $Z$. {\tt row/1} and
{\tt column/1} serve as ``type definitions'' for the arguments of the
open predicate position/2. The third component contains the
constraints. In principle, a constraint could be any first order logic
formula. However, my preference is for {\em rules}, universally
quantified implications. In my opinion, they offer the best compromise
between conciseness and readability\footnote{In this regard, I agree
  with the stable logic programming's use of rules to represent a unit
  of knowledge.}. A rule is a formula of the form $head \leftarrow
body$ with head a disjunction of literals (with ``;'' as separator)
and body a conjunction of literals (with ``,'' as separator. Because
it is common in PROLOG programs and to achieve more concise
formulations, we also use a disjunction of bodies in the position of a
body literal (as in the last constraint).  Similarly one could allow a
conjunction of heads in the position of a head literal; e.g.\ 
replacing the first two constraints by {\tt (row(R),column(C)) <-
  position(R,C)}. I write explicit {\tt true} for an empty conjunction
and {\tt false} for an empty disjunction to remind these are
constraints, not definitions or goals. The first two constraints
express the types of the arguments of the open predicate.  The next
two state that there is exactly one queen on every row.  Finally the
last constraint states that a queen should not attack a queen on a
later row i.e.\ should not be on the same column or diagonal. The
symmetry of the attack relation is used to avoid redundant constraints
(the presence of $R1<R2$).

It is fair to say that this formulation is more declarative than the
usual one in (Constraint) Logic Programming where a list of column
positions is built which represents a safe configuration.

Expressing the same problem as a {\bf Smodels} program, one obtains:
\begin{verbatim}
row(1..8) <- .
column(1..8) <- .
position(R,C) <- row(R), column(C), 
                 not negposition(R,C).
negposition(R,C) <- row(R), column(C), 
                    not position(R,C).
% at least one queen on each row
row_has_queen(R) <- row(R), column(C), 
                    position(R,C).
<- row(R), not row_has_queen(R).
% no more than one queen on each row
<- row(R), column(C1), column(C2), 
   position(R,C1), position(R,C2), C1\=C2.
% no two queens on same column
<- row(R1), row(R2), column(C), 
   position(R1,C), position(R2,C), R1\=R2.
% no two queens on same diagonal
<-  row(R1), row(R2), column(C1), column(C2),
    position(R1,C), position(R2,C), 
    R1\=R2, C1\=C2, abs(R2-R1)=abs(C2-C1).
\end{verbatim}
This program is a minor variant of the program in \cite{Nie00}. 

A first major difference with the abductive program is the pair of
rules ---typical for stable logic programming--- with {\tt position/2}
and {\tt negposition/2} in the head that defines the solution space of
the problem: for each candidate pair of coordinates {\tt (i,j)},
either {\tt position(i,j)} should be true ---there is a queen with
coordinates {\tt (i,j)}--- or {\tt negposition(i,j)} should be true
---there is no queen with coordinates {\tt (i,j)}---. In the abductive
program, the abduced {\tt position/2} facts are the definition of the
{\tt position/2} predicate, hence there is no queen with coordinates
{\tt (i,j)} when no such fact is abduced. On the other hand, the
abductive program has constraints to ensure that the abduced {\tt
  position/2} facts are well typed, i.e.\ have valid coordinates. A
second major difference is in the way the constraint that each row
must have a queen is expressed.  In the abductive program the
constraints {\tt row\_has\_queen(R) <- row(R)} and {\tt <- row(R), not
  row\_has\_queen(R)} have the same meaning. Both reject solutions for
$\Delta$ such that the unique model of $D \cup \Delta$ do not contain
{\tt row(n)} and {\tt row\_has\_queen(n)} for every {\tt n} that is a
row of the board.  However, in the stable model program, the
constraint has to be expressed as {\tt <- row(R), not
  row\_has\_queen(R)} i.e.\ a model which contains {\tt row(n)} but
not {\tt row\_has\_queen(n)} is rejected as a candidate stable model.
The constraint {\tt row\_has\_queen(R) <- row(R)} expresses that a
candidate stable model containing {\tt row(n)} is extended with {\tt
  row\_has\_queen(n)} which does not establish that there is a {\tt
  position(n,...)} atom in the model. This is a subtlety of stable
logic programming which is non-trivial for beginners. Perhaps the
difficulty for the beginning stable logic programmer lies in the
problem of recognising that the same notation ---rules--- are being
used for three purposes. In a rule with a head such as {\tt
  row\_has\_queen(R) <- row(R), column(C), position(R,C)}, the rule
defines the {\tt row\_has\_queen/1} predicate\footnote{To my
  understanding, it is almost as an inductive definition stating what
  is true about the predicate, but without bothering about what is
  false; if that matters, it has to be stated explicitly in other
  rules.}. A rule with an empty head such as {\tt <- row(R), not
  row\_has\_queen(R)} is a constraint, eliminating a number of
candidate stable models. Finally, rules with heads but which are not
inductive definitions, due to the loop through negation (as in the
rule pair with heads {\tt position(R,C)} and {\tt negposition(R,C)})
serve to generate a solution space for the predicate of interest (a
negative predicate has to be introduced to make the negative
information about the predicate of interest explicit).

A minor difference is that {\bf Smodels} accepts only domain
restricted programs, i.e.\ that each variable occurs in a positive
``domain'' predicate ({\tt row/1} or {\tt column/1}). This ensures
efficient handling of the so called grounding problem \cite{Nie00}.
Introducing some extra syntax for defining domains and for typing
predicates, e.g.\  
\begin{verbatim}
constant size == 8.
domain row == 1..size.
domain column == 1..size.
predicate position(row,column).
predicate negposition(row,column).
predicate row_has_queen(row).
\end{verbatim}
it is likely these extra calls could be automatically inserted. In
fact, also the abductive program could be simplified when including
domain declarations:
\begin{verbatim}
constant size == 8.
domain row == 1..size.
domain column == 1..size.
abducible position(row,column).
\end{verbatim}
The definitions of {\tt row/1} and {\tt column/1 } and the first two
constraints enforcing the well-typedness of abduced {\tt position/2}
facts can then be dropped.

\section{Planning}
\label{sec:planning}

\cite{Lif-ICLP99} describes the use of answer set programming for
planning problems. He is building on the work of other authors and we
refer to his paper for the broader context of this approach to
planning and the contributions of others in its development. A plan is
described by a history or evolution of a system over a fixed time
interval. In the concrete case of the blocks world, a configuration of
the world at a time point $T$ is described by a number of facts {\tt
  on(B,L,T)} expressing that block $B$ is on $L$ (another block or the
table) at time $T$. Evolution is caused by {\tt move(B,L,T)} actions
expressing that block $B$ is moved to location $L$ at time $T$. The
answer set programming approach consists of formulating constraints
such that the stable models (describing the successive configurations
and the move actions) only describe legal evolutions from some initial
configuration to some final configuration. Our intention is to show
that such problem can equally well be formulated in abductive logic
programming. The main problem discussed in \cite{Lif-ICLP99} is a
blocks world problem in a blocks world with two grippers. The problem
is to find a plan which converts an initial configuration into a final
configuration. In developing our solution, we try to follow the answer
set program (ASP) of Lifschitz as close as possible.

In what follows, the constant {\tt maxt} is used to represent the time
at which the final configuration must be reached. The ASP starts with
defining the ``solution space'' by the rules
\begin{equation}
  on(B,L,0); \neg on(B,L,0) \leftarrow 
\end{equation}
\begin{equation}
  move(B,L,T); \neg move(B,L,T) \leftarrow (T \neq maxt)
\end{equation}
As there are other rules ``defining'' {\tt on/3} for non-zero time
points, we isolate the initial configuration into an abducible {\tt
  initially\_on/2}, hence we use two abducibles, {\tt initially\_on/2}
and {\tt move/3}. To ensure the abduced facts are well typed, we
introduce integrity constraints such as {\tt movetime(T) <-
  move(B,L,T)} where {\tt movetime/1} is defined to be any time point
in the interval 0 to $maxt-1$.

Next we define {\tt on/3}.
\begin{equation}
  on(B,L,0) :-\ initially\_on(B,L)
\end{equation}
\begin{equation}
  on(B,L,T+1) :-\ move(B,L,T). \label{eq:move}
\end{equation}
\begin{displaymath}
  on(B,L,T+1) :-\ on(B,L,T), \\
\end{displaymath}
\begin{equation}
              \hspace{8em} not(terminates\_on(B,T))\ \label{eq:frame}
\end{equation}
The first clause defines {\tt on/3} for time point 0.  The second
clause (\ref{eq:move}) expresses that moving a block to a location
causes that block to be at that location in the next time point. The
third clause (\ref{eq:frame}) is the frame axiom, it expresses that
blocks stay in place at time $T+1$ if it was there at time $T$ and its
stay was not terminated at time $T$. In addition we have to define
{\tt terminates\_on/2}. It can be defined as
\begin{displaymath}
  terminates\_on(B,T) :-\  move(B,L1,T), 
\end{displaymath}
\begin{equation}
 \hspace{9em}   on(B,L0,T), L0\neq L1\ \
\end{equation}

The ASP also has the rule \ref{eq:move} for describing the effect of a
move. For the frame axiom it uses:
\begin{displaymath}
  on(B,L,T+1) \leftarrow  on(B,L,T), 
\end{displaymath}
\begin{equation}
  \hspace{7em}  not\ \neg on(B,L,T+1)  (t \neq maxt)
\end{equation}
This rule expresses the default that $B$ is still on $L$ at time $T+1$
if it was there at time $T$ and one fails to prove that it is not on
$L$ at time $T+1$\footnote{It is one of the rules which I found hard
  to grasp when learning about answer set programming.}. Because there
is now a rule which has $not\ \neg on(B,L,T+1)$ as a condition, i.e.\
the absence of a false fact, the ASP needs rules defining these false
facts. It uses:
\begin{equation}
  \neg on(B,L,T) \leftarrow on(B,L1,T), L \neq L1. \label{eq:2loc}
\end{equation}
Our decision to define {\tt on/3} has forced us to deviate from the
ASP. In my opinion it makes the program easier to understand. By the
way, our approach is also feasible with stable models. Niemel\"a uses a
rule similar to ours in his stable logic program for solving planning
problems \cite{Nie00}.

Our definition of {\tt on/3} is but a variant of the well known
acyclic PROLOG program solving the Yale Shooting Problem. It is a
widespread belief that it relies on negation as failure. For acyclic
programs \cite{Apt91}, many semantics coincide, in particular the
completion semantics coincides with the well-founded or inductive
definition semantics. As already stated before, our view is that such
programs make use of classical negation with respect to the implicit
higher order theory circumscribing the unique model.

The rest of the program consists of constraints expressing the
assumptions about the executability of {\tt move/3} steps. We have
the following constraints (see the program below for the actual code)
which also occur in the answer set program unless otherwise stated.
\begin{itemize}
\item A block cannot be moved if another block is on top of it.
\item A block cannot be moved to a moving target. (This subsumes the
  constraint that a block cannot be moved to itself).
\item A block cannot be moved to two locations. In the answer set
  program, this constraint is subsumed by the constraint (\ref{eq:2loc})
  that a block cannot be at two locations.
\item No two blocks can be on top of the same block.
\item The robot can perform only two moves at a time which we
  express by the constraint that one cannot move three blocks at a
  time (because we already excluded to move the same block to two
  different locations). The answer set program uses (without real
  need?) the more complex constraint that there cannot be three moves
  at the same time.
\item The answer set program also expresses the constraint that each
  block must be supported (directly or indirectly) by the table. It is
  easy to show that this property holds if it holds in the initial
  configuration (the target of a move is a block which is supported)
  or in the final configuration, hence we omit it. It would be needed
  if both initial and final configuration are incomplete. Likely the
  same argument holds for the answer set program.
\item Constraints expressing the initial and the final configuration.
\end{itemize}

The complete abductive program, including an initial and final
configuration, is as follows:

\begin{example}{A planning problem}
\begin{verbatim}

abducible(move/3). 
abducible(initially_on/2)

% DEFINITIONS

maxtime(3).
block(B) :- B in 1..6.
location(table).
location(L) :- block(L).
movetime(T) :- maxtime(T1), T0=T1-1, 
               T in 0..T0.

% definition of on/3
on(B,L,0) :- initially_on(B,L). 
% a block is at l at t+1 if moved to l at t
on(B,L,T+1) :- move(B,L,T).
% frame axiom (uses negation)
on(B,L,T+1) :- on(B,L,T)), 
               not(terminates_on(B,T)).
% auxiliary definition
terminates_on(B,T) :- move(B,L1,T), 
                      on(B,L0,T), L0 \= L1.

% CONSTRAINTS

% initial configuration
initially_on(1,2) <- true. 
initially_on(2,table)<- true.
initially_on(3,4) <- true.
initially_on(4,table) <- true.
initially_on(5,6) <- true.
initially_on(6,table) <- true.

% initially_on/2 is correctly typed
block(B) <- initially_on(B,L).
location(L) <- initially_on(B,L).

% move/3 is correctly typed
block(B) <- move(B,L,T).
location(L) <- move(B,L,T).
movetime(T) <- move(B,L,T).

% it is illegal to move a block if 
%       another block is on it 
false <- move(B,L,T), on(B1,B,T). 

% it is illegal to move a block to a moving 
%target (implies it cannot be moved to itself)
false <- move(B1,B2,T), move(B2,L2,T).

% a block cannot be moved to two locations
false <- move(B,L1,T), move(B,L2,T), L1 \= L2.

% no two blocks can be on top of same block, 
false <- on(B1,B,T), on(B2,B,T), 
         B1 \= B2, block(B).

% it is illegal to move three blocks 
%       at the same time
false <- move(B1,L1,T),move(B2,L2,T),
         move(B3,L3,T), 
         B1 \= B2, B1 \= B3, B2 \= B3.

% the final configuration
on(1,table,3) <- true. 
on(2,1,3) <- true.   
on(3,2,3)  <- true.
on(4,table,3)  <- true.  
on(5,4,3) <- true.  
on(6,5,3) <- true.

\end{verbatim}
\end{example}

I believe it is fair to claim that this program is from knowledge
representation point of view as good as the program given by
Lifschitz. 

\section{Conclusion}
\label{sec:conclussion}

In this paper we have repeated the viewpoint of Marc Denecker
\cite{Marc98} that good logic programs are inductive definitions and
that the use of negation in normal logic programs is better not
considered as negation as failure, but as classical negation with
respect to the higher order logic theory which circumscribes
the unique model of the program read as an inductive definition. A
correct program divides its world (the Herbrand base of the program)
in two halves, the true atoms and the false atoms. This logic is
non-monotonic, adding new clauses to the program may cause
non-monotonic changes to the unique model.

To cope with partial knowledge, we use a simple scheme for abduction,
where a set of abducibles $A$ identifies the predicates one is unable
to define completely with an inductive definition and where integrity
constraints $I$ are expressing the partial knowledge about the
abducibles. In this paradigm, the purpose is not to compute an answer
to the variables in a query, but to find a model for the abducibles
satisfying the integrity constraints\footnote{ A query {\tt ?- p(X)}
  can be translated in an integrity constraint {\tt false <- p(X),
    x(X)} with {\tt x/1} a new abducible, hence queries can still be
  supported}. Representing the model as a set of ground facts
$\Delta$, the computational task is to find a $\Delta$ such that the
integrity constraints $I$ are true in the unique model of the logic
program $P \cup \Delta$ (which should be a correct inductive
definition). It is currently unclear to me whether the requirement
that $P \cup \Delta$ is a correct inductive definition puts a
limitation on the usability of this paradigm. It is also unclear
whether there is need for the {\em minimality} \cite{Kakas98} of
abductive solutions $\Delta$.

With this form of negation, there is the problem that the negative
information is not recursively enumerable. Hence, it is impossible to
define a complete proof procedure. The answer to this is that there is
a need to prove termination of the programs one write (for the proof
strategy being used). In fact this is not different from the current
situation as proof procedures are also incomplete in practice for
other semantics. Perhaps the paradigm also creates a need for a
methodology and for (automated) methods to verify that a program is a
correct inductive definition. Partial solutions exist, for example
acyclic programs \cite{Apt91} are correct inductive definitions. Note
also that terminating proof procedures are feasible in the function
free case which is addressed by the stable logic programming paradigm.

Through two examples, the n-queens problem and a planning problem, we
have shown that our paradigm offers solutions to problems which are,
from knowledge representation point of view, as good as their
counterparts in the stable logic programming paradigm.

Personally, we believe that our approach even offers some advantages
(though we realise this is a matter of taste and of familiarity with
various formalisms). Our paradigm is a natural extension to logic
programming. We only add abducibles and integrity constraints to it
(and use a notation for the latter which distinguishes them from
program clauses defining predicates). Stable logic programming is a
rather radical departure from standard logic programming. Moreover we
see a number of drawbacks in its programming paradigm (which may well
be due to our ignorance and lack of familiarity and which can possibly
be solved by offering a more high level language). We observed that
the rule concept is used for different purposes: for defining what we
call the abducibles and their solutions space (by rules which fail to
be inductive definitions), for defining other predicates (by rules
which are inductive definitions) and for expressing integrity
constraints (by rules without head). Moreover there seems to be
sometimes a need for explicitly defining what is false (e.g. the rule
(\ref{eq:2loc}) in the planning problem; such rule is not required for the
predicate {\tt row\_has\_queen/1} in the queens problem. My
understanding is that such rules are needed when the negated predicate
is used as a default ($ not \ \neg \ldots$) in the condition of a
rule. This seems to spoil somewhat the modularity of the represented
knowledge. 

We have not discussed implementation. With some minor transformations,
our programs are executable under the abductive proof procedure SLDNFA
\cite{SLDNFA}. Recently this procedure has been integrated with a
finite domain constraint solver (SLDNFAC) \cite{Bert99}. SLDNFAC is a
Prolog meta interpreter which collects finite domain constraints and
passes them on to the finite domain constraint solver. It is inspired
by the work of Kakas on the ACLP system \cite{Kakas97,Kakas2000}.
Notwithstanding the overhead of meta interpretation it outperforms
{\bf Smodels} \cite{NS96,Nie00} on the n-queens problem \cite{PMDB00}.
On other problems (graph colouring), the performance is comparable.
In fact also other implementation paths are feasible, including
translation to {\bf Smodels}. E.g.\ \cite{abd:tms} describes how to
transform an abductive problem in a stable logic program.

Technically speaking, this paper offers little that is new. Inductive
definitions are being promoted by Marc Denecker for quite a while
\cite{Marc98}. Abduction has been widely studied and surveyed
\cite{Kakas,Kakas98}. If we added anything it was adding limitations:
insistence on the simple view that $P \cup \Delta$ should be an
inductive definition in which the integrity constraints are true.
While it is unclear whether this is expressive enough for all
purposes, hopefully it offers the advantage that it can be understood
with little effort by logic programmers unfamiliar with abduction and
stable model programming. For example by practitioners doing finite
domain CLP programming and which are now writing ``procedural''
programs for solving their problems using various versions of Prolog
extended with a finite domain solver. The language we propose enhanced
with some mechanisms to declare the finite domains over which the
abducibles range could perhaps be a useful language for specifying finite
domain CLP problems at a higher level than is currently the case.

Before us, other have proposed new languages incorporating abduction
for solving the class of problems we addressed. The n-queens problem
is also described in e.g.\ \cite{KWT98}. The use of abduction for
solving planning problems goes back to \cite{Eshghi88}.

\section{Acknowledgements}

I am grateful to the reviewers for their useful comments on the draft;
also to Marc Denecker, Nikolay Pelov and Bob Kowalski for various
discussions and useful comments. This research is supported by the GOA
project LP+; the author is supported by FWO-Vlaanderen.

\end{document}